\documentclass{article}

\usepackage{scicite}

\usepackage[utf8]{inputenc} 
\usepackage[T1]{fontenc}    
\usepackage{hyperref}       
\usepackage{url}            
\usepackage{booktabs}       
\usepackage{amsfonts}       
\usepackage{nicefrac}       
\usepackage{microtype}      

\usepackage{amsmath}
\usepackage{graphicx}
\usepackage{xcolor}
\usepackage{booktabs}
\usepackage{algorithm}
\usepackage{algorithmic}
\usepackage{xspace}
\usepackage{xcolor}
\usepackage{multirow}
\usepackage{svg}
\usepackage{tabularx}
\usepackage{graphicx}

\newcommand{\match}{game\xspace}
\newcommand{\matches}{games\xspace}
\newcommand{\round}{round\xspace}
\newcommand{\rounds}{rounds\xspace}

\newcommand{\yaku}{{yaku}\xspace}

\newcommand{\dora}{{dora}\xspace}

\title{Suphx: Mastering Mahjong with Deep Reinforcement Learning\footnote{This work was conducted at Microsoft Research Asia. The 2nd, 4th, 5th, and 6th authors were interns at Microsoft Research Asia then.}}
\date{}

\author
{
    Junjie Li,$^1$ Sotetsu Koyamada,$^2$ Qiwei Ye,$^1$\\ 
    Guoqing Liu,$^3$ Chao Wang,$^4$ Ruihan Yang,$^5$ Li Zhao,$^1$\\   Tao Qin,$^1$ Tie-Yan Liu,$^1$ Hsiao-Wuen Hon$^1$\\
\\
\\
\normalsize{$^{1}$Microsoft Research Asia}
\\
\normalsize{$^{2}$Kyoto University}
\\
\normalsize{$^{3}$University of Science and Technology of China}
\\
\normalsize{$^{4}$Tsinghua University}
\\
\normalsize{$^{5}$Nankai University}
}

\begin{document}

\maketitle

\begin{abstract}
Artificial Intelligence (AI) has achieved great success in many domains, and game AI is widely regarded as its beachhead since the dawn of AI. In recent years, studies on game AI have gradually evolved from relatively simple environments (e.g., perfect-information games such as Go, chess, shogi or two-player imperfect-information games such as heads-up Texas hold'em) to more complex ones (e.g., multi-player imperfect-information games such as multi-player Texas hold'em and StartCraft II). Mahjong is a popular multi-player imperfect-information game worldwide but very challenging for AI research due to its complex playing/scoring rules and rich hidden information. We design an AI for Mahjong, named Suphx, based on deep reinforcement learning with some newly introduced techniques including global reward prediction, oracle guiding, and run-time policy adaptation.  Suphx has demonstrated stronger performance than most top human players in terms of stable rank and is rated above 99.99\% of all the officially ranked human players in the Tenhou platform. This is the first time that a computer program outperforms most top human players in Mahjong. 
\end{abstract}
\section{Introduction}
\label{sec:intro}
Building superhuman programs for games is a long-standing goal of artificial intelligence (AI). Game AI has made great progress in past two decades ~\cite{Tesauro1995-za,Silver2016-ws,Silver2017-gx,Silver2018-rs,Bowling2017-tm,Moravcik2017-cf,Brown2018-wg}. Recent studies have gradually evolved from relatively simple perfect-information or two-player games (e.g., shogi, chess, Go, and heads-up Texas hold'em) to more complicated imperfect-information multi-player ones (e.g., contract bridge~\cite{rong2019competitive}, Dota~\cite{berner2019dota}, StarCraft II~\cite{vinyals2019grandmaster}, and multi-player Texas hold'em~\cite{brown2019superhuman}). 

Mahjong, a multi-round tile-based game with imperfect information and multiple players, is very popular with hundreds of millions of players worldwide. In each round of a Mahjong game, four players compete with each other towards the first completion of a winning hand. Building a strong Mahjong program raises great challenges to the current studies on game AI. 

First, Mahjong has complicated scoring rules. Each \match of Mahjong contains multiple \rounds, and the final ranking (and so the reward) of a \match is determined by the accumulated round scores of those \rounds. The loss of one \round does not always mean that a player plays poorly for that \round (e.g., the player may tactically lose the last \round to ensure rank 1 of the \match if he/she has a big advantage in previous \rounds), and so we cannot directly use the \round score as a feedback signal for learning. Furthermore, Mahjong has a huge number of possible winning hands. Those winning hands can be very different from each other, and different hands result in different winning round scores of the round. Such scoring rules are much more complex than previously studied games including chess, Go, etc. A professional player needs to carefully choose what kind of winning hand to form in order to trade off the winning probability and winning score of the round.

Second, in Mahjong each player has up to 13 private tiles in his/her hand which are not visible to other players, and there are 14 titles in the dead wall, which are invisible to all the players throughout the game,  and 70 tiles in the live wall, which will become visible once they are drawn and discarded by the players. As a result, on average, for each information set (a decision point of a player), there are more than $10^{48}$ hidden states that are indistinguishable to him/her. Such a large set of hidden information makes Mahjong a much more difficult imperfect-information game than previously studied ones such as Texas hold'em poker. It is hard for a Mahjong player to determine which action is good only based on his/her own private tiles, since the goodness of an action highly depends on the private tiles of other players and the wall tiles that are invisible to everyone. Consequently, it is also difficult for an AI to connect the reward signal to the observed information. 

Third, the playing rule of Mahjong is complex: (1) there are different types of actions including Riichi, Chow, Pong, Kong, discard, and (2) the regular order of plays can be interrupted when making a meld (Pong or Kong), going Mahjong (declaring a winning hand), or robbing a Kong. Because each player can have up to 13 private tiles, it is hard to predict those interruptions, and therefore we even cannot build a regular game tree; even if we build a game tree, such a tree will have a huge number of paths between the consecutive actions of a player.   This prevents the direct application of previously successful techniques for games such as Monte-Carlo tree search~\cite{silver2016mastering,Silver2018-rs} and counterfactual regret minimization~\cite{Brown2018-wg,brown2019superhuman}.

Due to the above challenges, although there are several attempts~\cite{Mizukami2015-vm,Kurita2019-cd,Gao2019-wl,Village_undated-jt}, the best Mahjong AI is still far behind top human players.

In this work, we build \emph{Suphx} (short for Super Phoenix), an AI system for 4-player Japanese Mahjong (Riichi Mahjong), which has one of the largest Mahjong communities in the world.  Suphx adopts deep convolutional neural networks as its models. The networks are first trained through supervised learning from the logs of human professional players and then boosted through self-play reinforcement learning (RL), with the networks as the policy. We use the popular policy gradient algorithm \cite{sutton2000policy} for self-play RL and introduce several techniques to address the aforementioned challenges.

\begin{enumerate}
    \item Global reward prediction trains a predictor to predict the final reward (after several future \rounds) of a \match based on the information of the current and previous \rounds. This predictor provides effective learning signals so that the training of the policy network can be performed. In addition, we design look-ahead features to encode the rich possibilities of different winning hands and their winning scores of the round, as a support to the decision making of our RL agent.

\item Oracle guiding introduces an oracle agent that can see the perfect information including the private tiles of other players and the wall tiles. This oracle agent is a super strong Mahjong AI due to the (unfair) perfect information access. In our RL training process, we gradually drop the perfect information from the oracle agent, and finally convert it to a normal agent which only takes observable information as input.  With the help of the oracle agent, our normal agent improves much faster than standard RL training which only leverages observable information. 

     \item  As the complex playing rules of Mahjong lead to an irregular game tree and prevent the application of Monte-Carlo tree search techniques, we introduce parametric Monte-Carlo policy adaptation (pMCPA) to improve the run-time performance of our agent. pMCPA gradually modifies and adapts the offline-trained policy to a specific \round in the online playing stage when the \round goes on and there is more information observable (such as public tiles discarded by four players). 
 
\end{enumerate}

We evaluated Suphx on the most popular and competitive Mahjong platform, Tenhou~\cite{Tsunoda_undated-fr}, which has more than 350,000 active users.
Suphx reached 10 dan at Tenhou, and its stable rank, which describes the long-term average of the performance of a player, surpasses most top human players.

\section{Overview of Suphx} \label{sec:suphx}
In this section, we first describe the decision flow of Suphx, and then the network structures and features used in Suphx.

\subsection{Decision Flow}
\label{sec:flow}
Due to the complex playing rules of Mahjong, Suphx learns five models to handle different situations: the discard model, the Riichi model, the Chow model, the Pong model, and the Kong model, as summarized in Table \ref{tab:5model}.
\begin{table}[]
    \centering
    \begin{tabular}{c|c}
    \hline \hline
    Model & Functionality \\ \hline
       Discard model  & Decide which tile to discard in normal situations \\
        Riichi model & Decide whether to declare Riichi\\
      Chow model & Decide whether/what to make a Chow\\
    Pong model & Decide whether to make a Pong \\
    Kong model & Decide whether to make a Kong\\     \hline\hline
    \end{tabular}
    \caption{Five models in Suphx}
    \label{tab:5model}
\end{table}

Besides these five learned models, Suphx employs another rule-based winning model to decide whether to declare a winning hand and win the \round. It basically checks whether a winning hand can be formed from a tile discarded by other players or drawn from the wall, and then makes decisions according to the following simple rules:
\begin{itemize}
    \item If this is not the last \round of the \match, declare and win the \round;
    \item If this is the last \round of the \match, 
    \begin{itemize}
        \item If after declaring a winning hand the accumulated round score of the whole \match is the lowest among the four players, do not declare;
        \item Otherwise, declare and win the \round.
    \end{itemize}
\end{itemize}

\begin{figure}[h!]
    \centering
    \includegraphics[width=1.0\textwidth]{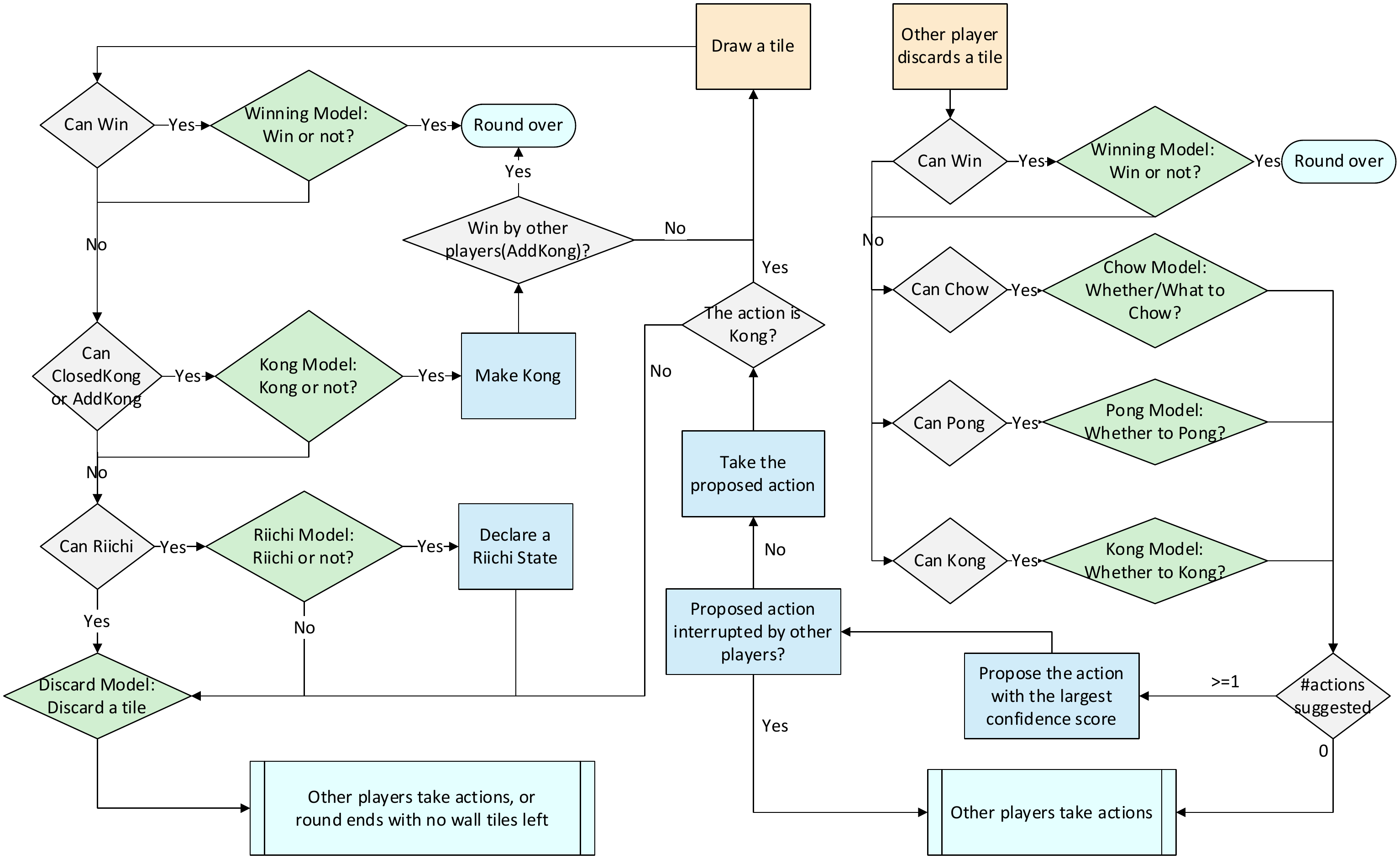}
    \caption{{Decision flow of Suphx.}
        }
    \label{fig:suphx}
\end{figure}

There are two kinds of situations that a Mahjong player needs to take actions, so does our AI Suphx (see Figure \ref{fig:suphx}):
\begin{itemize}
    \item The Draw situation: Suphx draws a tile from the wall. If its private tiles can form a winning hand with the drawn tile, the winning model decides whether to declare winning. If yes, it declares and the round is over. Otherwise, 
    \begin{enumerate}
        \item Kong step: If the private tiles can make an ClosedKong or AddKong with the drawn tile, the Kong model decides whether to make the ClosedKong or AddKong. If not, go to the Riichi step; otherwise, there are two sub cases:
        \begin{enumerate}
            \item If it is an ClosedKong, make the ClosedKong and go back to the Draw situation.
            \item If it is an AddKong, other players may win the round using this AddKong tile. If some other player wins, the round is over; otherwise, make the AddKong and go back to the Draw situation.
        \end{enumerate}
        \item Riichi step: If the private tiles can make Riichi with the drawn tile, the Riichi model decides whether to declare Riichi. If not, go to the Discard step; otherwise, declare Riichi and then go to the Discard step. 
        \item Discard step: The discard model chooses a tile to discard. After that, it is the other players' turn to take actions or the round ends with no wall tiles left.\footnote{If a \round ends with no wall tiles left, this is also known as an exhaustive draw. 
        }
    \end{enumerate}
    \item The Other-Discard situation: Other players discard a tile. If Suphx can form a winning hand with this tile, the winning model decides whether to declare winning. If yes, it declares and the round is over. Otherwise, it checks whether a Chow, Pong, or Kong can be made with the discarded tile. If not, it is the other players' turn to take actions; otherwise, the Chow, Pong, or Kong model decides what action to take:
    \begin{enumerate}
        \item If no action is suggested by the three models, it is the turn for other players to take actions or the round ends with no wall tiles left. 
        \item If one or more actions are suggested, Suphx proposes the action with the highest confidence score (outputted by those models). If the proposed action is not interrupted by higher-priority actions from other players, Suphx takes the action, and then goes to the Discard step in the first situation. Otherwise, the proposed action is interrupted and it becomes the turn for other players to take actions. 
    \end{enumerate}
\end{itemize}

\subsection{Features and Model Structures}
Since deep convolutional neural networks (CNNs) have demonstrated powerful representation capability and been verified in playing games like chess, shogi and Go, Suphx also adopts deep CNNs as the model architecture for its policy.  

\begin{figure}[h!]
    \centering
    \includegraphics[width=11cm]{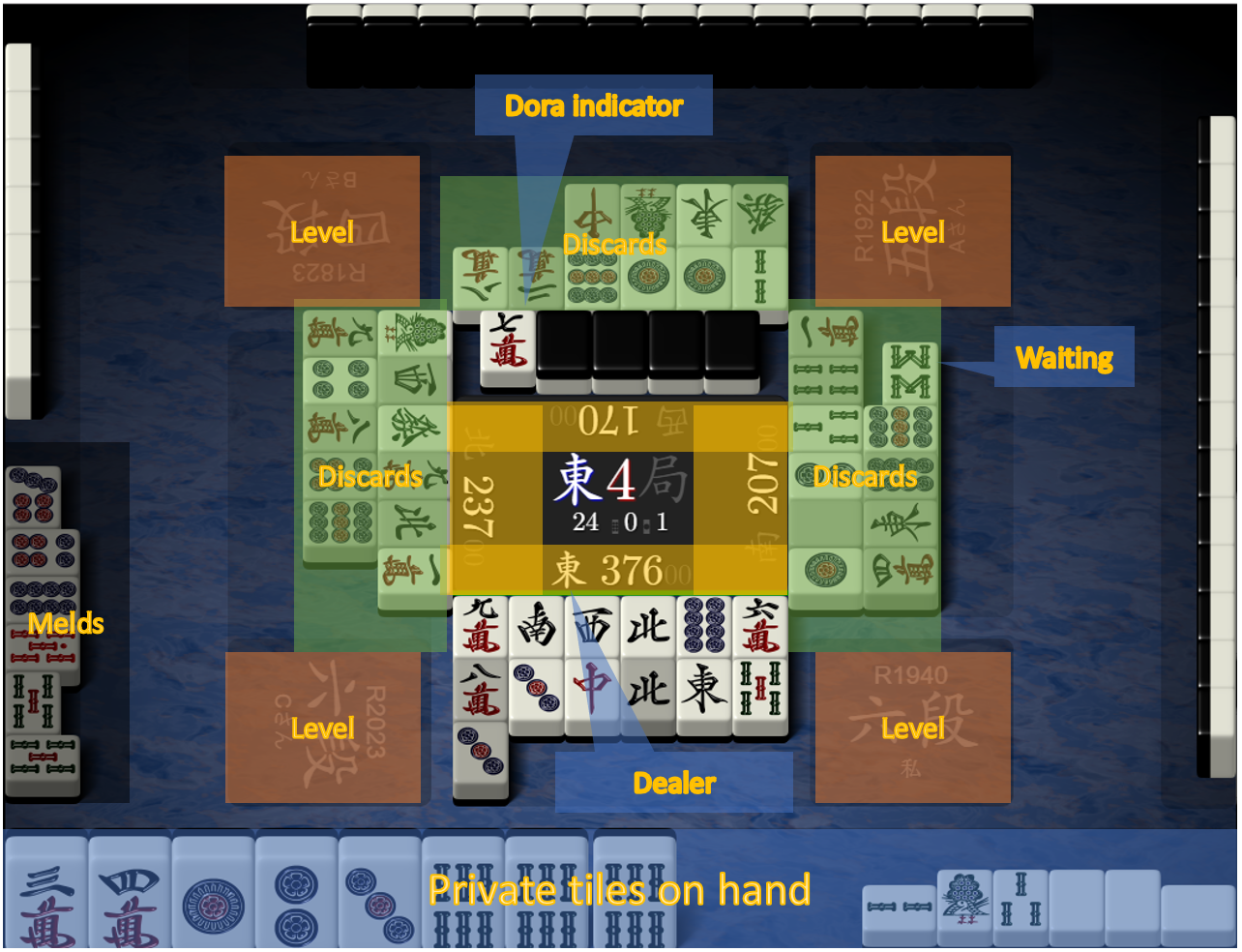}
    \caption{{Example of a state.} 
    Mahjong's state contains several types of information:
    (1) the {tile set} including private tiles, open hand, and doras,
    (2) the sequence of discarded tiles,
    (3) integer features including accumulated round scores of the four players and the number of tiles left in the live wall, and
    (4) {categorical} features including round id, dealer, counters of repeat dealer, and Riichi bets.
    }
    \label{fig:mahjong-state-example}
\end{figure}

Different from board games like Go and Chess, the information (as shown in Figure \ref{fig:mahjong-state-example}) available to players in Mahjong is not naturally in the format of images. We carefully design a set of features to encode the observed information into channels that can be digested by CNNs.

\begin{figure}[h!]
    \centering
    \includegraphics[width=1.0\textwidth]{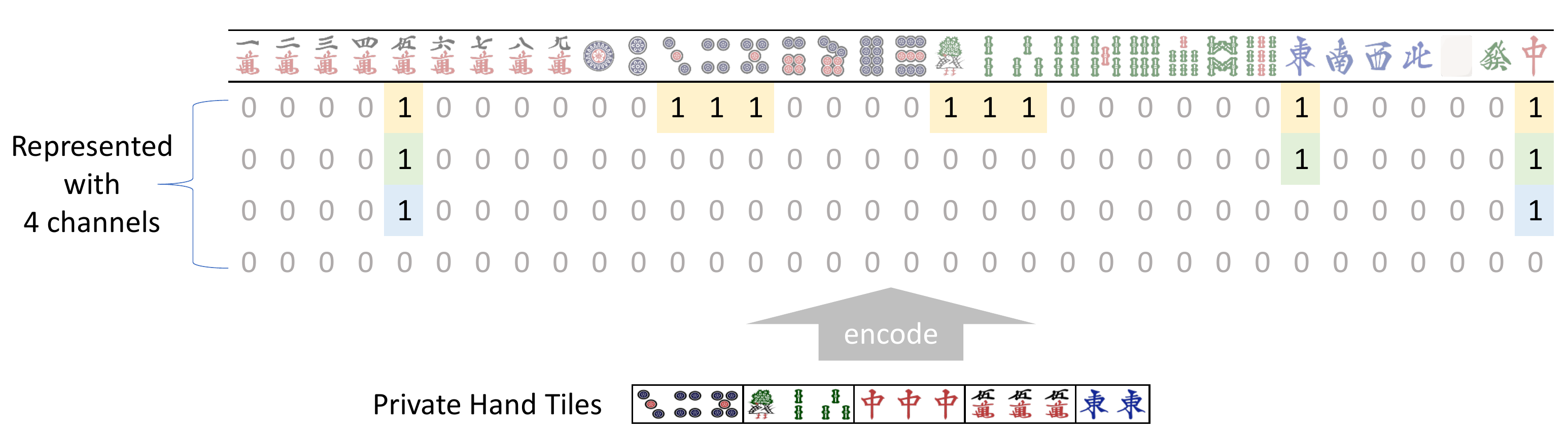}
    \caption{{Encoding of private tiles.}
    We encode the private tiles of a player into four channels.
    There are four rows and 34 columns, with each row corresponding to one channel, and each column indicating one type of tile. 
    The \textit{m}-th column in the \textit{n}-th channel means 
        whether there are \textit{n} tiles of the \textit{m}-th type in the hand.
    }
    \label{fig:hand-example-representation}
\end{figure}

As there are 34 unique tiles in Japanese Mahong, we use multiple $34 \times 1$ channels to represent a state. As shown in Fig. \ref{fig:hand-example-representation}, we use four channels to encode the private tiles of a player. Open hand, doras, and the sequence of discarded tiles are encoded similarly into other channels. Categorical features are encoded into multiple channels, with each channel being  either all 0's or all 1's. Integer features are partitioned into buckets and each bucket is encoded using a channel of either all 0's or all 1's.

In addition to the directly observable information, we also design some look-ahead features, which indicate the probability and round score of winning a hand if we discard a specific tile from the current hand tiles and then draw tiles from the wall to replace some other hand tiles. In Japanese Mahjong, a winning hand of 14 tiles contains four melds and one pair. There are 89 kinds of melds and 34 kinds of pairs, which lead to a huge number of different possible winning hands. Furthermore, different hands result in different winning scores of the \round according to the complex scoring rules.\footnote{A common list of different scoring patterns could be found at \url{http://arcturus.su/wiki/List_of_yaku}} It is impossible to enumerate all the combinations of different discarding/drawing behaviors and winning hands. Therefore, to reduce computational complexity, we make several simplifications while extracting look-ahead features: (1) We perform depth first search to find possible winning hands. (2) We ignore opponents' behaviors and only consider drawing and discarding behaviors of our own agent.    With those simplifications, we obtain 100+ look-ahead features, with each feature corresponding to a 34-dimensional vector. For example, a feature represents whether discarding a specific tile can lead to a winning hand of 12,000 round score with replacing 3 hand tiles by tiles drawn from the wall or discarded by other players.

\begin{figure}[h!]
    \centering
    \includegraphics[width=1.0\textwidth]{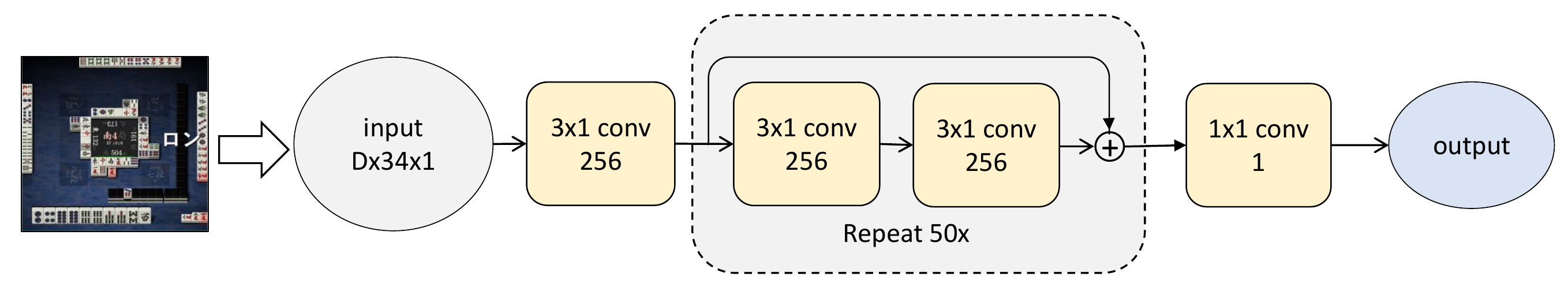}
    \caption{{Structure of the discard model}}
    \label{fig:discard-network}
\end{figure}

\begin{figure}[h!]
    \centering
    \includegraphics[width=1.0\textwidth]{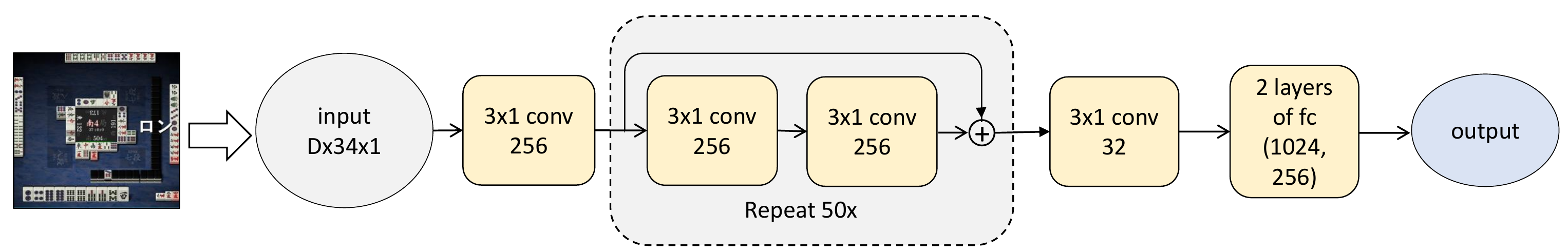}
    \caption{{Structures of the Riichi, Chow, Pong, and Kong models}}
    \label{fig:binary-network}
\end{figure}

In Suphx, all the models (i.e., the discard/Riichi/Chow/Pong/Kong models) use similar network structures (Figures \ref{fig:discard-network} and \ref{fig:binary-network}), except the dimensions of the input and output layers (Table \ref{tab:dimension}). The discard model has 34 output neurons corresponding to 34 unique tiles, the Richii/Chow/Pong/Kong models have only two output neurons corresponding to whether or not to take a certain action. In addition to the state information and look-ahead features, the inputs of Chow/Pong/Kong models also contain information about what tiles to make a Chow/Pong/Kong. Note that there is no pooling layer in our models, because every column of a channel has its semantic meaning and pooling will lead to information loss. 

\begin{table}[h!]
    \centering
    \begin{tabular}{c|c|c|c|c|c}
    \hline\hline
   &  Discard  & Riichi & Chow & Pong & Kong\\
    \hline
    Input & \multicolumn{2}{|c|}{$ 34 \times 838$}& \multicolumn{3}{|c}{$ 34 \times 958$}\\ \hline
    Output &34& \multicolumn{4}{|c}{2}\\ 
    \hline\hline
   \end{tabular}
   \caption{Input/out dimensions of different models}
   \label{tab:dimension}
\end{table}

\section{Learning Algorithm}

The learning of Suphx contains three major steps. First, we train the five models of Suphx by supervised learning, using  (state, action) pairs of top human players collected from the Tenhou platform. 
Second, we improve the supervised models through self-play reinforcement learning (RL), with the models as policy. We adopt the popular policy gradient algorithm (Section \ref{sec:pipeline}) and introduce global reward prediction (Section \ref{sec:reward}) and oracle guiding (Section \ref{sec:oracle}) to handle the unique challenges of Mahjong. Third, during online playing, we employ run-time policy adaptation (Section \ref{sec:adapt}) to leverage the new observations on the current round in order to perform even better.

\subsection{Distributed Reinforcement Learning with Entropy Regularization} 
\label{sec:pipeline}

The training of Suphx is based on distributed reinforcement learning.  In particular, we employ the policy gradient method and leverage importance sampling to handle the staleness of trajectories due to asynchronous distributed training:

\begin{equation}
\label{eq:obj}
   \mathcal{L}\left(\theta\right)=\underset{s, a \sim \pi_{\theta^{\prime}}}{\mathrm{E}}\left[\frac{\pi_{\theta}(a | s)}{\pi_{\theta^{\prime}}(a | s)} A^{\pi_{\theta}}(s, a)\right],
\end{equation}
where $\theta^{\prime}$ is (the parameters of) an old policy generating trajectories for training, $\theta$ is the latest policy to update, and $A^{\pi_{\theta}}(s, a)$ is the advantage of action $a$ at state $s$ with respect to policy $\pi_\theta$.

We find that RL training is sensitive to the entropy of the policy.  If the entropy is too small, RL training converges quickly and self-play does not significantly improve the policy; if the entropy is too large, RL training becomes unstable and the learned policy is of large variance. Therefore, we regularize the entropy of policy during the RL training as follows:
\begin{align}
\nabla_{\theta} J\left(\pi_{\theta}\right)=\underset{s,a \sim \pi_{\theta'}}{\mathrm{E}}\left[\frac{\pi_\theta(s,a)}{\pi_{\theta'}(s,a)} \nabla_{\theta} \log \pi_{\theta}\left(a | s \right) A^{\pi_{\theta}}\left(s, a\right)\right] + \alpha\nabla_{\theta}H(\pi_{\theta})
\end{align}
where $H(\pi_{\theta})$ is the entropy of policy $\pi_\theta$ and $\alpha>0$ is a trade-off coefficient. To ensure stable exploration, we dynamically adjust $\alpha$ to increase/decrease the entropy term if the entropy of our policy is smaller/larger than the target ${H}_{{\rm target}}$ in a recent period: 
\begin{align}
  \alpha \leftarrow \alpha + \beta \bigl( {H}_{{\rm target}} - \bar{H}(\pi_{\theta})\bigr),
\end{align}
where $\bar{H}(\pi_{\theta})$ is the empirical entropy of the trajectories in a recent period, and $\beta>0$ is a small step size.

\begin{figure*}[!t]\label{fig::overview}
	\centering
    \includegraphics[width=1.0\linewidth]{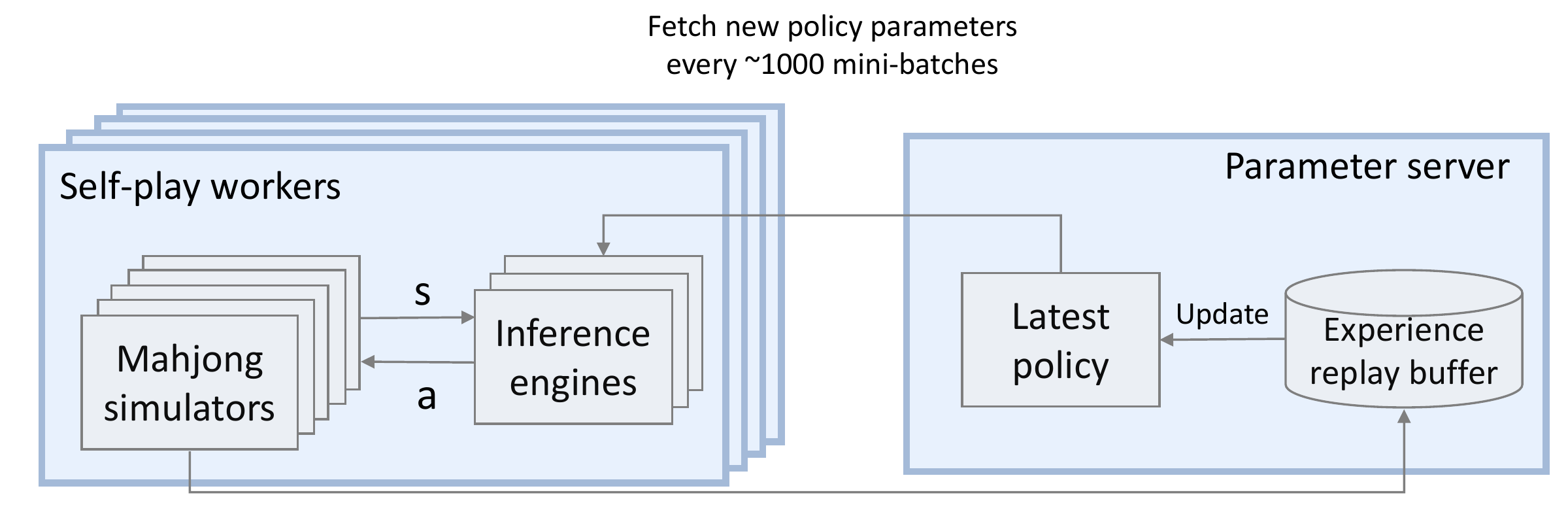} 
    \caption{\label{fig:systemoverview} Distributed RL system in Suphx
    }
\end{figure*}

The distributed RL system used by Suphx is shown in Figure \ref{fig:systemoverview}. The system consists of multiple self-play workers, each containing a set of CPU-based Mahjong simulators  and a set of GPU-based inference engines to generate trajectories. The update of the policy $\pi_\theta$ is decoupled from the generation of the trajectories:  a parameter server is used  to update the policy using multiple GPUs based on a replay buffer. During training, every Mahjong simulator randomly initializes a game with our RL agent as a player and other three opponents. When any of the four players needs to take an action, the simulator sends the current state (represented by a feature vector) to a GPU inference engine, which then returns an action to the simulator. GPU inference engines pull the up-to-date policy $\pi_\theta$ from the parameter server in a regular basis, to ensure that the self-play policy is close enough to the latest policy $\pi_\theta$.

\subsection{Global Reward Prediction}
\label{sec:reward}

In Mahjong, each \match contains multiple \rounds, e.g., 8-12 \rounds in Tenhou.\footnote{The number of \rounds of a \match is not fixed, depending on the win/loss of each \round. More details can be found at \url{https://tenhou.net/man/}.}
A \round starts with 13 private tiles dealt to each player,  in turn players draw and discard tiles, the \round ends until one of the players completes a winning hand or no tile is left in the wall, and then each player gets a {\round score}.
For example, the player who forms a winning hand gets a positive round score, and the others get zero or negative round scores. When all the \rounds end, each player gets a \match reward based on the rank of the accumulated \round scores.  

Players receive \round scores at the end of each \round, and receive \match rewards after 8-12 \rounds. However, neither \round scores nor \match rewards are good signals for RL training:
\begin{itemize}
    \item Since multiple \rounds in the same \match share the same \match reward, using \match rewards as a feedback signal cannot differentiate well-played \rounds and poorly-played \rounds. Therefore, one should better measure the performance of each \round separately.
    \item While the \round score is computed for each individual round, it may not be able to reflect the goodness of the actions, especially for top professional players. For example, in the last one or two \rounds of a \match, the rank-1 player with a big lead in terms of accumulated \round scores will usually become more conservative, and may purposely let rank-3 or rank-4 players win this \round so that he/she can safely keep rank 1 overall. That is, a negative \round score may not necessarily mean a poor policy: it may sometimes reflect certain tactics and thus correspond to a fairly good policy.
\end{itemize}

\begin{figure}[h!]
    \centering
    \includegraphics[width=0.70\textwidth]{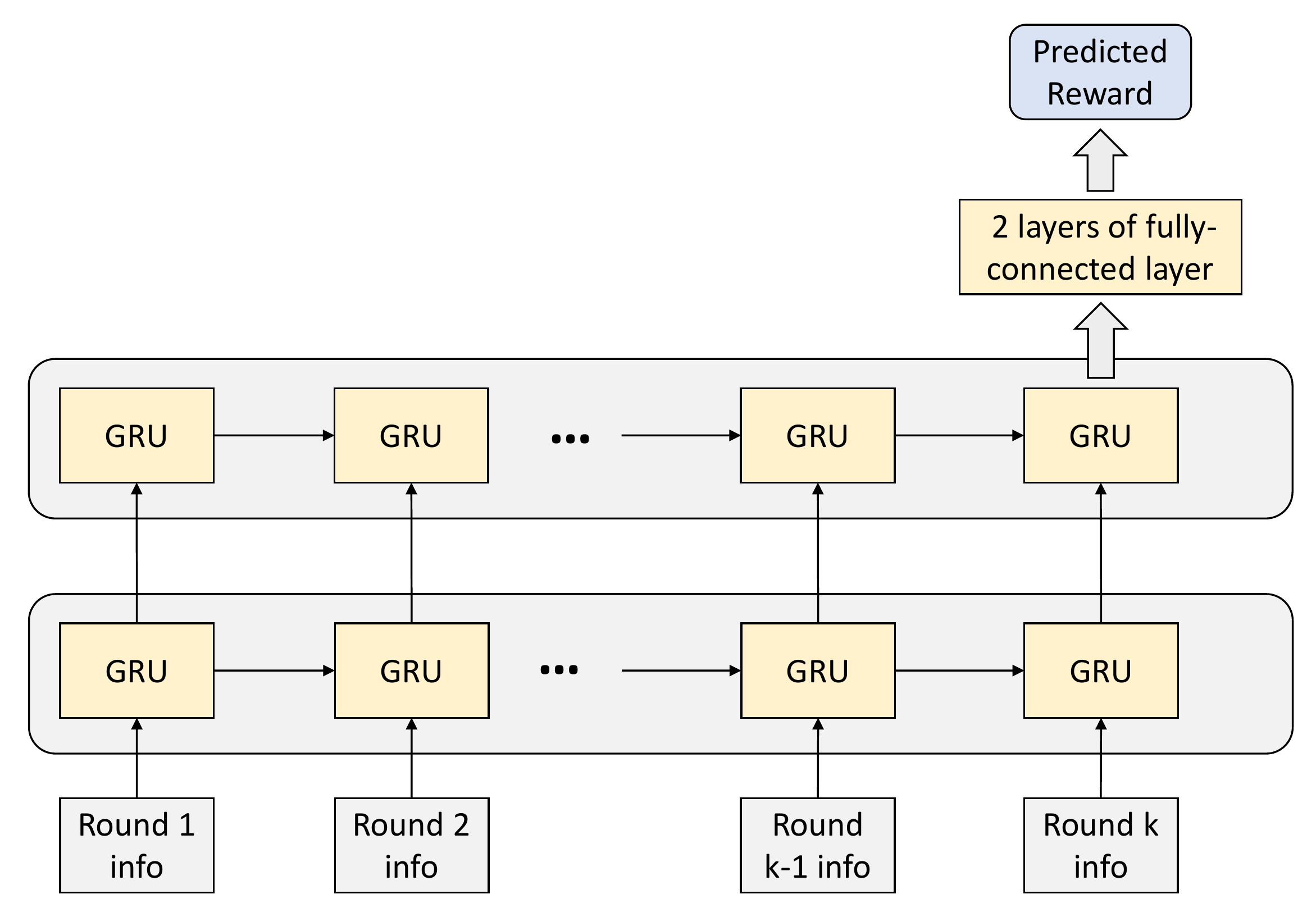}
    \caption{{Reward predictor: GRU network}}
    \label{fig:rnn-network}
\end{figure}

Therefore, to provide effective signal for RL training, we need to appropriately attribute the final \match reward (a global reward) to each \round of the \match. For this purpose, we introduce a global reward predictor $\Phi$, which predicts the final \match reward given the information of the current \round and all previous \rounds of this \match. In Suphx, the reward predictor $\Phi$ is a recurrent neural network, more specifically, a two-layer gated recurrent unit (GRU) followed by two fully-connected layers, as shown in Figure \ref{fig:rnn-network}. 

The training data for this reward predictor $\Phi$ come from the logs of top human players in Tenhou, and $\Phi$ is trained by minimizing the following mean square error:
\begin{equation}
    \label{eq:reward}
    \min\frac{1}{N}\sum_{i=1}^N\frac{1}{K_i}\sum_{j=1}^{K_i} (\Phi(x_i^1,\cdots, x_i^j)-R_i)^2,
\end{equation}
where $N$ denotes the number of games in the training data, $R_i$ denotes the final game reward of the $i$-th game, $K_i$ denotes the number of \rounds in the $i$-th game, $x_i^k$ denotes the feature vector of the $k$-th \round in the $i$-th game, including the score of this \round, the current accumulated round score, the dealer position, the counters of repeat dealer and Riichi bets.

When $\Phi$ is well trained, for a self-play game with $K$ \rounds, we use $\Phi(x^k)-\Phi(x^{k-1})$ as the reward of the $k$-th \round for RL training.

\subsection{Oracle Guiding}
\label{sec:oracle}
There is rich hidden information in Mahjong (e.g., the private tiles of other players and the wall tiles). Without access to such hidden information, it is hard to take good actions. This is a fundamental reason why Mahjong is a difficult game. In this situation, although an agent can learn a policy by means of reinforcement learning, the learning could be very slow. To speed up the RL training, we introduce an oracle agent, which can see all the perfect information about a state: (1) private tiles of the player, (2) open (previously discarded) tiles of all the players, (3) other public information such as the accumulated \round scores and Riichi bets, (4) private tiles of the other three players, and (5) the tiles in the wall. Only (1)(2) and (3) are available to the normal agent,\footnote{``Normal'' here  means that the agent has no access to the perfect information.} while (4) and (5) are additional ``perfect" information that is only available to the oracle. 

With the (unfair) access to the perfect information, the oracle agent will easily become a master of Mahjong after RL training. Here the challenge is how to leverage the oracle agent to guide and accelerate the training of our normal agent. According to our study, simple knowledge distillation does not work well: it is difficult for a normal agent, who only has limited information access, to mimic the behavior of a well-trained oracle agent, who is super strong and far beyond the capacity of a normal agent. Therefore, we need a smarter way to guide our normal agent with the oracle. 

To this end, there might be different approaches. In Suphx, what we do is to first train the oracle agent through reinforcement learning, using all the features including the prefect ones. Then we gradually drop out the perfect features so that the oracle agent will eventually transit to a normal agent:
\begin{equation}
\label{eq:obj.oracle}
   \mathcal{L}\left(\theta\right)=\underset{s, a \sim \pi_{\theta^{\prime}}}{\mathrm{E}}\left[\frac{\pi_{\theta}(a |[x_n(s),\delta_t x_o(s)])}{\pi_{\theta^{\prime}}(a | [x_n(s),\delta_t x_o(s)])} A^{\pi_{\theta}}([x_n(s),\delta_t x_o(s)], a)\right],
\end{equation}
where  $x_n(s)$ denote the normal features and $x_o(s)$ the additional perfect features of state $s$, and $\delta_t$ is the dropout matrix at the $t$-th iteration whose elements are Bernoulli variables with $P(\delta_t(i,j)=1)=\gamma_t$. We gradually decay $\gamma_t$ from 1 to 0.  When $\gamma_t=0$, all the prefect features are dropped out and the model transits from the oracle agent to a normal agent. 

After $\gamma_t$ becomes zero, we continue the training of the normal agent for a certain number of iterations. We adopt two tricks during the continual training. First, we decay the learning rate to one tenth. Second, we reject some state-action pairs if the importance weight is larger than a pre-defined threshold. According to our experiments, without these tricks, the continual training is not stable and does not lead to further improvements. 

\subsection{Parametric Monte-Carlo Policy Adaptation}
\label{sec:adapt}
The strategy of a top human player will be very different when his/her initial hand (private tiles) varies. For example, he/she will play aggressively to win more given a good initial hand, and conservatively to lose less given a poor initial hand. This is very different from previous games including Go and StarCraft. Therefore, we believe that we could build a stronger Mahjong agent if we are able to adapt the offline-trained policy in the run time. 

Monte-Carlo tree search (MCTS) is a well established technique in games like Go \cite{silver2016mastering} for run-time performance improvement. Unfortunately, as aforementioned, the playing order of Mahjong is not fixed and it is hard to build a regular game tree. Therefore, MCTS cannot be directly applied to Mahjong. In this work, we design a new method, named parametric Monte-Carlo policy adaptation (pMCPA).

When a \round begins and the initial private hand is dealt to our agent, we adapt the offline-trained policy to this given initial hand as follows:
\begin{enumerate}
        \item Simulations: Randomly sample private tiles for the three opponents and wall tiles from the pool of tiles excluding our own private tiles, and then use the offline-trained policy to roll out and finish the whole trajectory. In total, $K$ trajectories are generated in this way.
    \item Adaptation: Perform gradient updates using the rollout trajectories to finetune the offline policy. 
    \item Inference: Use the finetuned policy to play against other players in this \round. 
\end{enumerate}

Let $h$ denote the private hand tiles of our agent at a \round, $\theta_o$ denote the parameters of the policy trained off-line, and $\theta_a$ the parameters of the new policy adapted to this \round. Then we have
\begin{equation}
\theta_a=\arg\max_\theta \sum_{\tau\sim_{\theta_o} \mathcal{T}(h)}R(\tau)\frac{p(\tau;\theta)}{p(\tau;\theta_o)},    
\end{equation}
where $\mathcal{T}(h)$ is the set of trajectories with prefix $h$, and $p(\tau;\theta)$ is the probability of policy $\theta$ generating trajectory $\tau$. 

    According to our study, the number $K$ of simulations/trajectories does not need to be very large and  pMCPA does not need to collect statistics for all the following states for this \round. Since pMCPA is a parametric method, the updated policy (using the $K$ simulations) can lead to updated estimation of those states not visited in the simulations as well.  That is, such run-time adaptation can help to generalize our knowledge obtained from limited simulations to unseen states.

Please note that the policy adaptation is performed for each \round independently. That is, after we adapt the policy of our agent in the current \round, for the next \round, we will restart from the offline-trained policy once again.

\section{Offline Evaluation}
In this section, we report the effectiveness of each technical component of Suphx through offline experiments. 

\subsection{Supervised Learning}
In Suphx, the five models were first trained through supervised learning separately. Each training sample is a state-action pair collected from human professional players with state serving as the input and action serving as the label for supervised learning. For example, for the training of the discard model, the input of a sample is all the observable information (and the look-head features) of a state, and the label is the action taken by a human player, i.e., the tile discarded at this state.

The training data sizes and the test accuracy are reported in Table \ref{tab:sl}. The sizes of the validation data and test data are 10K and 50K respectively, for all the models. Since the discard model addresses a 34-class classification problem, we collected more training samples for it. As can be seen from the table, we achieve an accuracy of $76.7\%$ for the discard model, $85.7\%$ for the Riichi model, $95.0\%$ for the Chow model, $91.9\%$ for the Pong model, and $94.0\%$ for the Kong model. We also list the accuracy achieved by previous works~\cite{gao2018supervised} as a reference.\footnote{We would like to point out that our numbers are not directly comparable to previous numbers due to different training/test data and model structures.}

\begin{table}[]
    \centering
    \begin{tabular}{cccc}
    \hline \hline
    Model & Training Data Size &  Test Accuracy & Previous Accuracy~\cite{gao2018supervised}\\ \hline
       Discard model  & 15M &   76.7\% & 68.8 \%\\
        Riichi model & 5M &   85.7 \% &-\\
      Chow model & 10M &  95.0\% & 90.4\%\\
    Pong model & 10M &   91.9 \% & 88.2\%\\
    Kong model & 4M &  94.0 \% &-\\     \hline\hline
    \end{tabular}
    \caption{Results for supervised learning}
    \label{tab:sl}
\end{table}

\subsection{Reinforcement Learning}

To demonstrate the value of each RL component in Suphx, we trained several Mahjong agents:
\begin{itemize}
    \item SL: the supervised learning agent. This agent (with all the five models) was trained in a supervised way, as illustrated in the previous subsection.
    \item SL-weak: an under-trained version of the SL agent, which serves as opponent models while evaluating other agents.
    \item RL-basic: the basic version of the reinforcement learning agent. In RL-basic, the discard model was initialized with the SL discard model and then boosted through the policy gradient method with \round scores as reward and entropy regularization. The Riichi, Chow, Pong, and Kong models remain the same as those of the SL agent.\footnote{These four models can also be improved through reinforcement learning, although not as significant as the discard model. Therefore, to reduce training time, we inherit the SL models.} 
    \item RL-1: the RL agent that enhances RL-basic with global reward prediction. The reward predictor was trained through supervised learning with human game logs from Tenhou.
    \item RL-2: the RL agent that further enhances RL-1 with oracle guiding. Please note that in both RL-1 and RL-2, we also only trained the discard model using RL, and left the other four models the same as those of the SL agent. 
\end{itemize}

The initial private tiles have large randomness and will greatly impact the win/loss of a \match. To reduce the variance caused by initial private tiles, during the offline evaluation, we randomly generated one million \matches. Each agent plays against 3 SL-weak agents on these \matches. In such a setting, the evaluation of one agent took 20 Tesla K80 GPUs for two days. For the evaluation metric, we computed the stable rank of an agent following the rules of Tenhou (see Appendix C). To reduce the variance of stable rank, for each agent, we randomly sampled 800K \matches from the one million \matches, for 1000 times.

\begin{figure}[h!]
  \begin{center}
    \includegraphics[width=10cm]{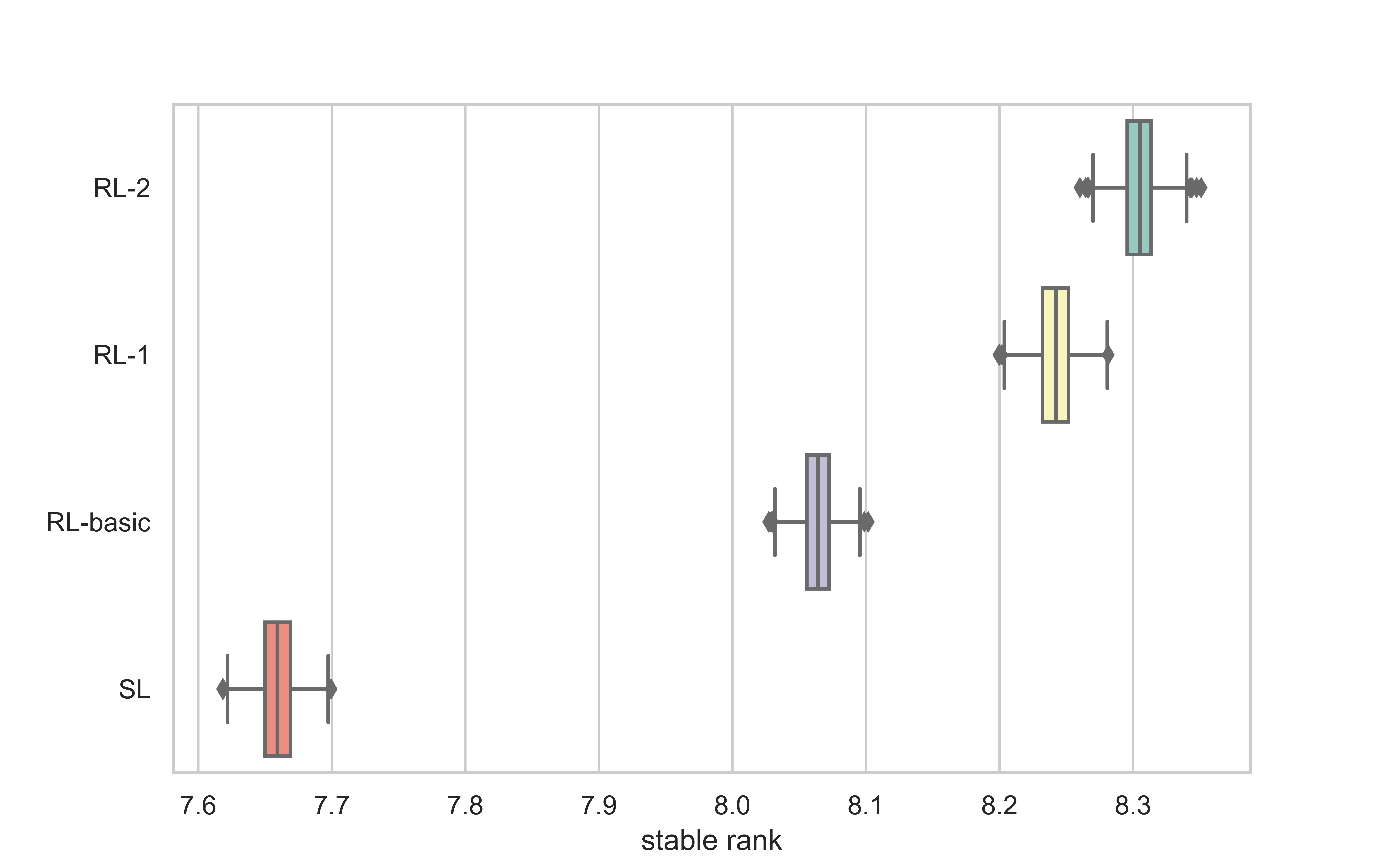}
  \end{center}
  \caption{{Statistics of stable rank over one million \matches.} The  plot shows the three quartile values of a distribution along with extreme values. The “whiskers” extend to points that lie within 1.5 IQRs of the lower and upper quartile, and then observations that fall outside this range are displayed independently. }
\label{fig:offline}
\end{figure}

Figure \ref{fig:offline} shows the interquartile ranges\footnote{According to Wikipedia, "In descriptive statistics, the interquartile range (IQR) is a measure of statistical dispersion, being equal to the difference between 75-th and 25-th percentiles, or between upper and lower quartiles, IQR = Q3 -  Q1. In other words, the IQR is the first quartile subtracted from the third quartile; these quartiles can be clearly seen on a box plot on the data. It is a trimmed estimator, defined as the 25\% trimmed range, and is a commonly used robust measure of scale."} of stable ranks over the 1000 samplings for those agents. Note that for fair comparison, each RL agent was trained using 1.5 million \matches. The training of each agent costs 44 GPUs (4 Titan XP for the parameter server and 40 Tesla K80 for self-play workers) and two days. As can be seen, RL-basic leads to good improvement over SL, RL-1 outperforms RL-basic, and RL-2 brings additional gains against RL-1. These experimental results clearly demonstrate the value of  reinforcement learning, as well as the additional values of global reward prediction and oracle guiding. 

\begin{figure}[h!]
  \begin{center}
    \includegraphics[width=10cm]{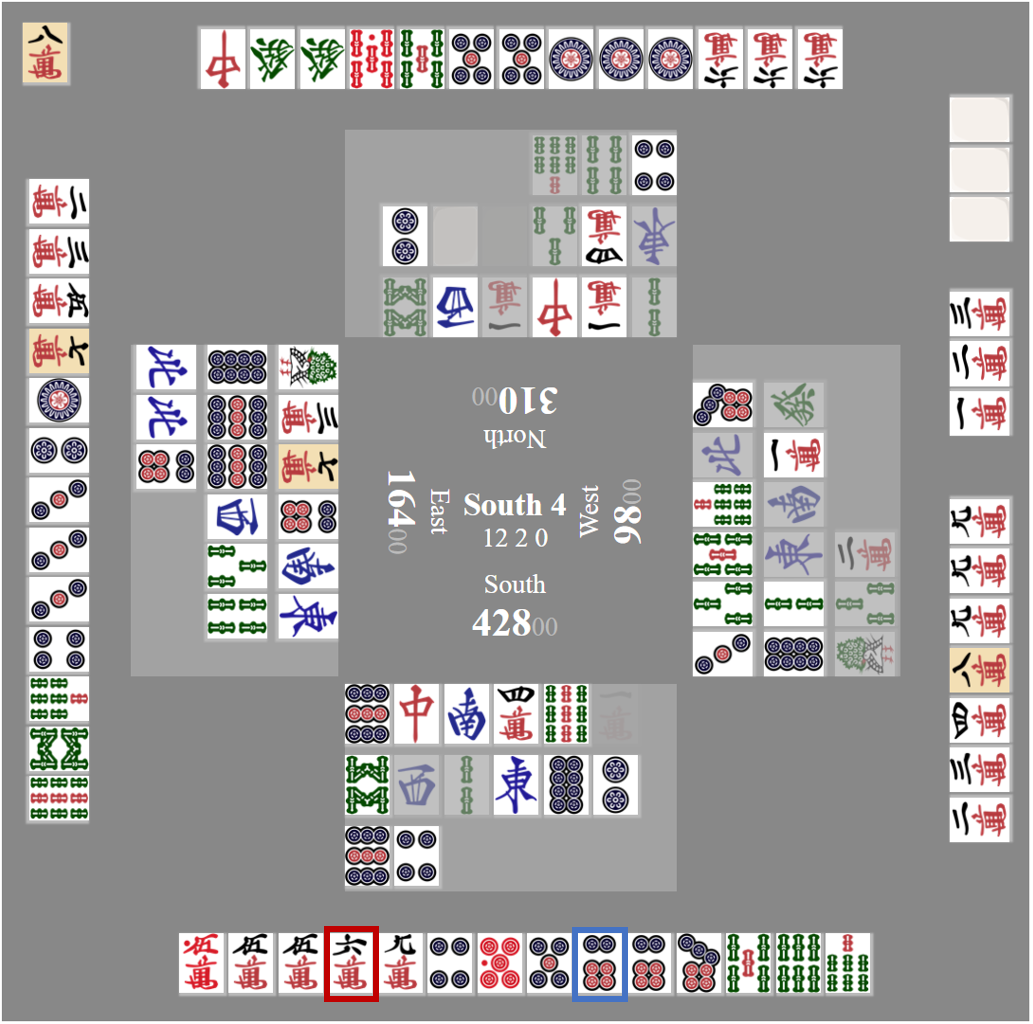}
  \end{center}
  \caption{{With global reward prediction, our agent (the south player) plays conservatively when its accumulated round score has a big lead in the last round of a game, even if its hand tiles are good and it has a certain probability to win this round. The RL-basic agent discards the red-boxed tile to win this round, but discarding this tile is risky since the same tile has not been discarded by any player in this round. In contrast, RL-1 and RL-2 agents play in a defense mode and discard the blue-boxed tile, which is a safe tile because the same tile has just been discarded by the west player.}}
\label{fig:folding-on-waiting}
\end{figure}

With the global reward predictor distributing the game reward to each \round, the trained agent can better maximize the final game reward instead of the \round score. For example, in Figure \ref{fig:folding-on-waiting}, our agent (the south player) has a big lead with good hand in the last round of the game. Based on the current accumulated \round scores of the four players, winning this round gets only marginal reward while losing this round will lead to a big punishment. Therefore instead of playing aggressively to win this round, our agent plays conservatively, chooses the safest tile to discard, and finally gets the first place/rank for this game. In contrast, RL-basic discards another tile to win the round, which brings a big risk of losing the 1-st rank of the entire game.

\subsection{Evaluation of Run-Time Policy Adaptation}

In addition to testing the enhancement to offline RL training, we also tested the run-time policy adaptation. The experimental setting is described as follows.

When a \round begins and the private tiles are dealt to our agent,
\begin{enumerate}
    \item Data generation:  We fix the hand tiles of our agent and simulate 100K trajectories. In each trajectory, the hand tiles of the other three players and wall tiles are randomly generated, and we use four copies of our agent to roll out and finish the trajectory.
    \item Policy adaptation: We fine-tune and update the policy trained offline over those 100K trajectories by using the basic policy gradient method.
    \item Test of the adapted policy: Our agent plays against the other three players using the updated policy on another 10K test set where the private tiles of our agent are still fixed. As the initial private tiles of our agent is fixed, the performance of the adapted agent on this test set can tell whether such run-time policy adaptation really makes our agent adapt and work better for the current private tiles.
\end{enumerate}

Please note that run-time policy adaptation is time consuming due to the roll-outs and online learning. Therefore, at the current stage, we only tested this technique on hundreds of initial \rounds. The wining rate of the adapted version of RL-2 against its non-adapted version is 66\%, which demonstrates the advantage of run-time policy adaptation. 

{}

\begin{figure}[h!]
  \begin{center}
    \includegraphics[width=10cm]{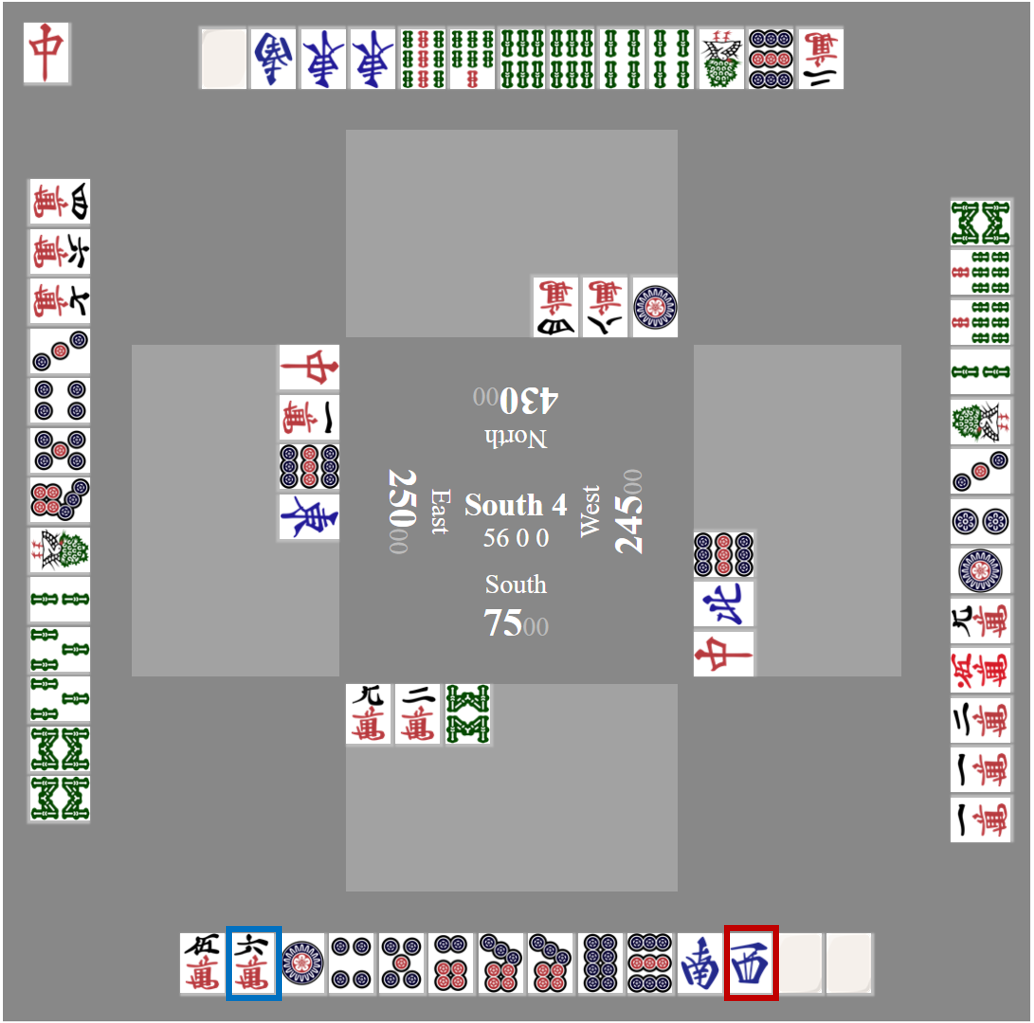}
  \end{center}
  \caption{{In this example, to move out of the 4-th place of the game, the agent needs to win more than 12,000 round score in this round. Through simulations, the agent learns to know that discarding the red-boxed tile is easy to win this round; however, the corresponding winning round score will be less than 12,000. After adaptation, the agent discards the blue-boxed tile, which leads to lower probability of winning but more than 12,000 winning round score once it wins. By doing so, it takes risk and successfully move out of the 4-th place.}}
  
\label{fig:adaption-case-study}
\end{figure}

Policy adaption makes our agent work better for the current private hand, especially at the last 1 or 2 rounds of a game. Figure \ref{fig:adaption-case-study} shows an example of the last round of a game. Through simulations the agent learns to know that while it is easy to win this round with a nice round score, this is unfortunately not enough to avoid ending the game with the 4-th place. Thus, after adaptation, the agent plays more aggressively, takes more risks, and eventually wins the round with a much larger round score and successfully avoid ending the game with the 4-th place.

\section{Online Evaluation}

To evaluate the real performance of our Mahjong AI Suphx\footnote{Suphx is equivalent to RL-2 trained with about 2.5 million games. Given that Tenhou.net has time constraint for each action, run-time policy adaptation was not integrated into Suphx while testing on Tenhou.net since it is time consuming. We believe the integration of run-time policy adaptation will further improve Suphx.}, we let it play on Tenhou.net, the most popular online platform for Japanese Mahjong. Tenhou has two major rooms, the expert room and the phoenix room. The expert room is open to AI and human players of 4 dan and above, while the phoenix room is only open to human players of 7+ dan. According to this policy, Suphx can only play in the expert room. 

Suphx played 5000+ games in the expert room and achieved 10 dan in terms of record rank\footnote{Record rank is the highest rank a player has ever achieved in Tenhou. As shown in Appendix B, the rank of a player is dynamic and usually changes over time. For example, if he/she does not play well recently, his/her rank will drop.} and 8.74 dan in terms of stable rank.\footnote{See Appendix C for the definition of stable rank.} It is the first and only AI in Tenhou that achieves 10 dan in terms of record rank. 

We compare Suphx with several AI/human players in Table~\ref{table:compared-other-AIs1}: 

\begin{itemize}
    \item Bakuuchi \cite{mizukami2015building}: This is a Mahjong AI designed by the University of Tokyo based on Monte Carlo simulation and opponent modeling. It does not use reinforcement learning. 
    \item NAGA\footnote{\url{https://dmv.nico/ja/articles/mahjong_ai_naga/}}: This is a Mahjong AI designed by Dwango Media Village based on deep convolutional neural networks. It does not use reinforcement learning either. 
    \item We also compare Suphx with top human players who have achieved 10 dan in terms of record rank. To be fair, we only compare their game playing in the expert room after they achieved 10 dan. Since these top human players spent the majority of their time in the phoenix room (partly due to its more friendly scoring rules) and only played occasionally in the expert room after they achieved 10 dan, we can hardly calculate a reliable stable rank for each individual of them.\footnote{We chose to compare with the performance of top human players according to their game playing in the expert room but not the phoenix room because these two rooms have different scoring rules and the stable ranks are not directly comparable.} Therefore, we treat them as one macro player to make a statistically reasonable comparison.
\end{itemize}

\begin{figure}[h!]
  \begin{center}
    \includegraphics[width=10cm]{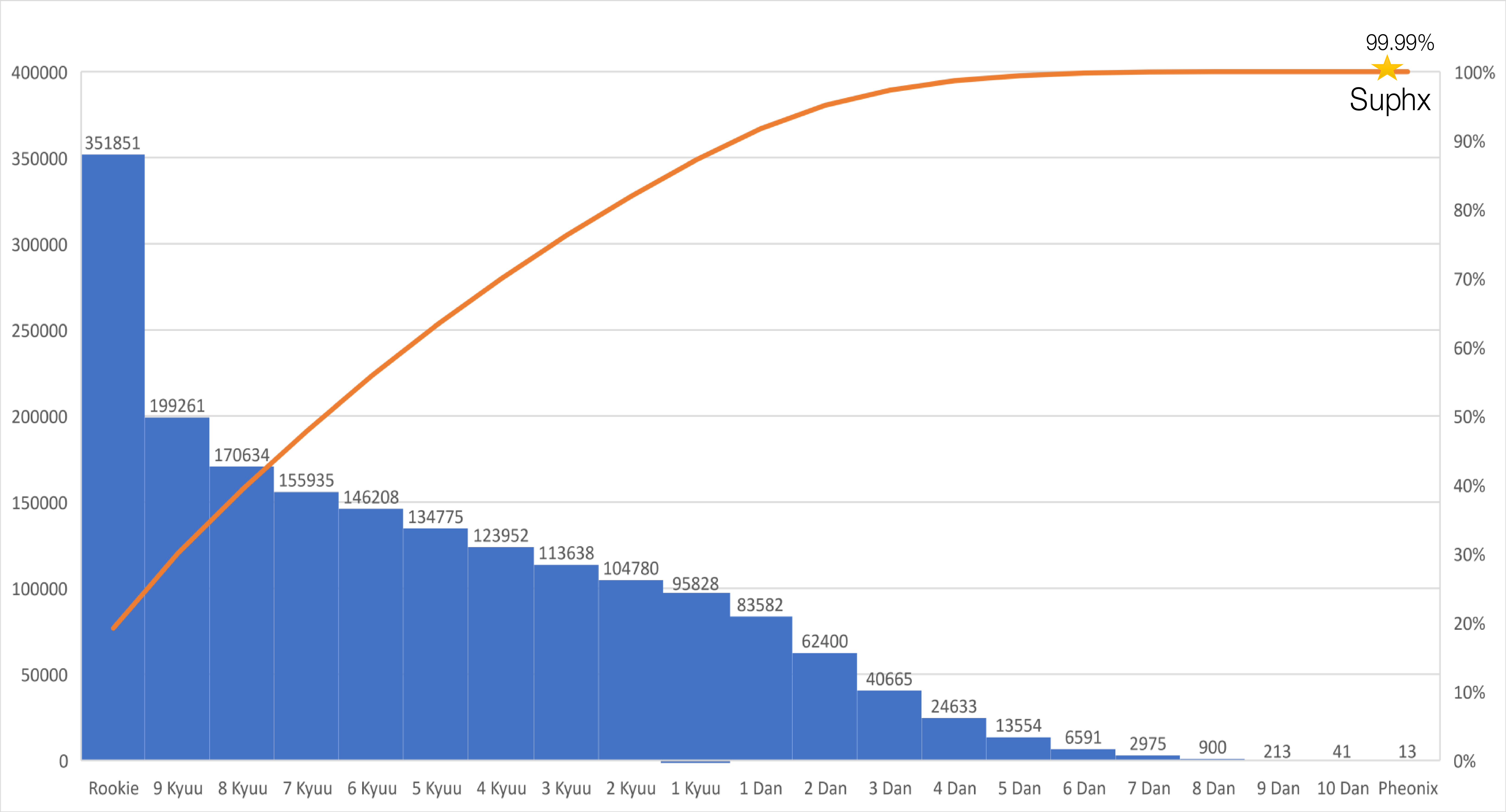}
  \end{center}
  \caption{{Distributions of record ranks of human players in Tenhou. Each bar indicates the number of human players above a certain level in Tenhou.}
  }
  \label{fig:rank}
\end{figure}

We can see that in terms of stable rank, Suphx is about 2 dan better than  Bakuuchi and NAGA, the two best Mahjong AIs before Suphx. Although these top human players have achieved the same record rank (10 dan) as Suphx, they are not as strong as Suphx in terms of stable rank. Figure \ref{fig:rank} plots the distributions of record ranks\footnote{\textit{Phoenix} is a honorable title in Tenhou, when a 10-dan player gets 4000 ranking points. There are only 13 players (and 14 accounts) in Tenhou's history won this honorable title for 4-player Mahjong.} of current active users in Tenhou, which shows that Suphx is above 99.99\% human players in Tenhou.

\begin{table}[h!]
  \begin{center}
    \begin{tabular}{cccc}
    \hline\hline
    & {\#Game} & {record rank} & {Stable Rank} \\
    \hline
      Bakuuchi & 30,516 & 9 dan & 6.59 \\
      NAGA   & 9,649           & 8 dan & 6.64 \\
  Top human    & 8,031           & 10 dan           & 7.46 \\
      Suphx    & 5,760           & 10 dan           & \textbf{8.74} \\
\hline\hline    \end{tabular}
  \end{center}
  \caption{{ Comparison with other AIs and top human players.}}
  \label{table:compared-other-AIs1}
\end{table}

\begin{figure}[h!]
  \begin{center}
    \includegraphics[width=12cm]{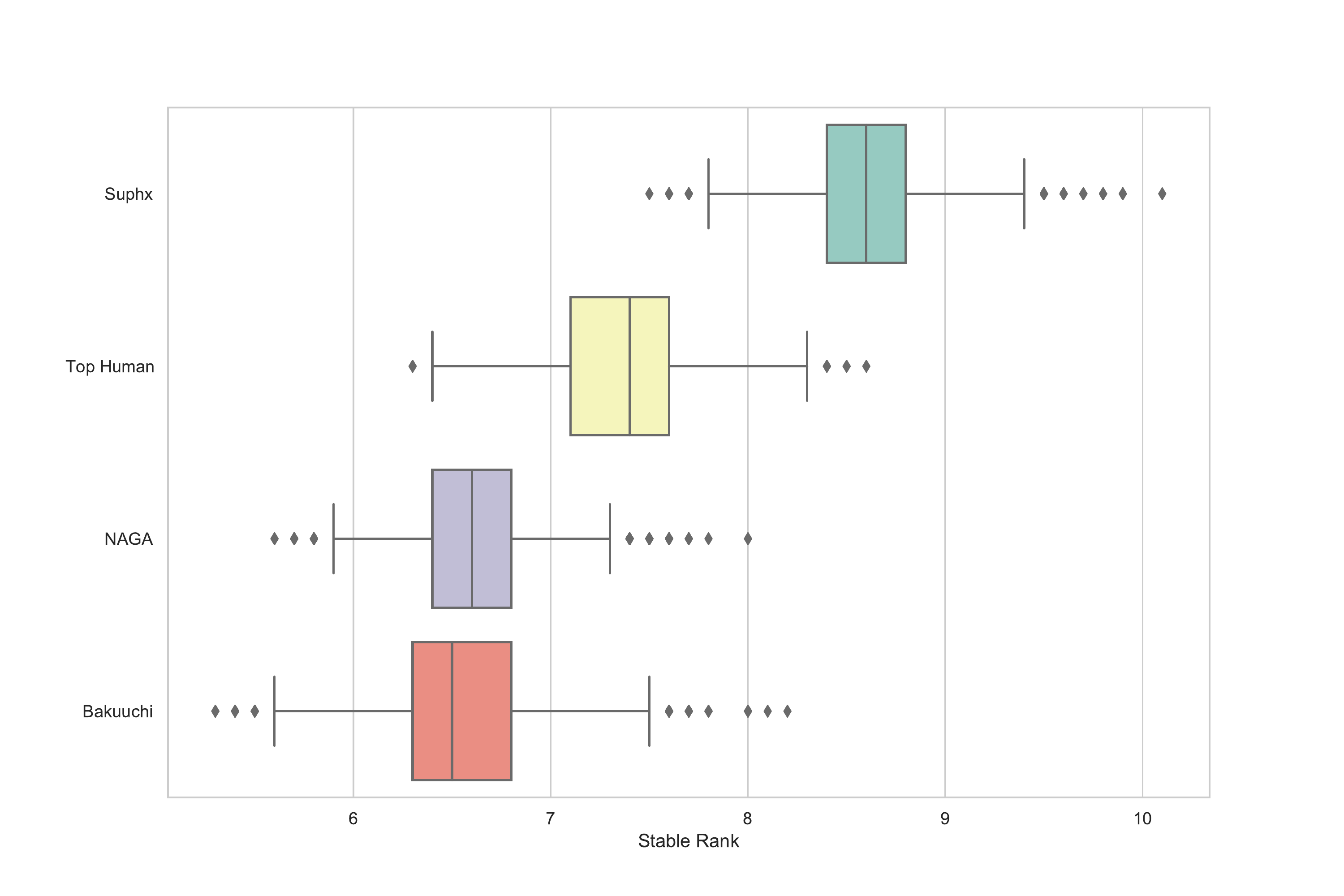}
  \end{center}
  \caption{{Statistics of stable ranks with $K=2000$ and $N=5000$.}
   }
  \label{fig:expert-room-performance}
\end{figure}

As discussed in Appendix B, the record rank sometimes cannot reflect the true level of a player: for example, there are 100+  players with a record rank of 10 dan in Tenhou's history , but their true levels can be very different. The stable rank is more stable (by its definition) and of finer granularity than the record rank; however, it could also be of large variance, especially when a player has not played enough number of games in Tenhou. Thus, in order to make more informative and reliable comparisons, we proceed as follows. For each AI/human player, we randomly sample $K$ games from its/his/her logs in the expert room and compute the stable rank using those $K$ games. We do such sampling for $N$ times and show the statistics of the corresponding $N$ stable ranks of each player in Figure \ref{fig:expert-room-performance}. As can be seen, Suphx surpasses both the other two AIs and the average performance of top human professional players with a big margin.

We further show more statistics of those AI/human players in Table \ref{tab:rankdis}. We have several interesting observations from the table:
\begin{itemize}
    \item Suphx is very strong at defense and has very low deal-in rate. This is confirmed by the comments from top human players on Suphx.\footnote{A quote from a \textit{pheonix} human player can be found at \url{https://twitter.com/Futokunaio_Sota/status/1142399895577325568}}
    \item Suphx has very low 4-th rank rate, which is the key to get a high stable rank in Tenhou according to its scoring rules.
\end{itemize}
\begin{table}[h!]
  \begin{center}
    \begin{tabularx}{\linewidth}{rXXXXXX}
   \hline\hline
    & 1st Rank & 2nd Rank & 3rd Rank & 4th Rank & Win Rate  & Deal-in Rate \\
    \hline
      Bakuuchi & 28.0\% & 26.2\% & 23.2\% & 22.4\% & 23.07\% & 12.16\% \\
      NAGA   & 25.6\% & 27.2\% & 25.9\% & 21.1\%  & 22.69\% & 11.42\% \\
  Top human    & 28.0\% & 26.8\% & 24.7\% & 20.5\% &-&- \\
        Suphx    & 29.3\% & 27.5\% & 24.4\% & \textbf{18.7\%} & 22.83\% & \textbf{10.06\%} \\
    \hline\hline
    \end{tabularx}
  \end{center}
  \caption{{More statistics: rank distribution and win/deal-in rate}}
  \label{tab:rankdis}
\end{table}

\begin{figure}[h!]
  \begin{center}
    \includegraphics[width=12cm]{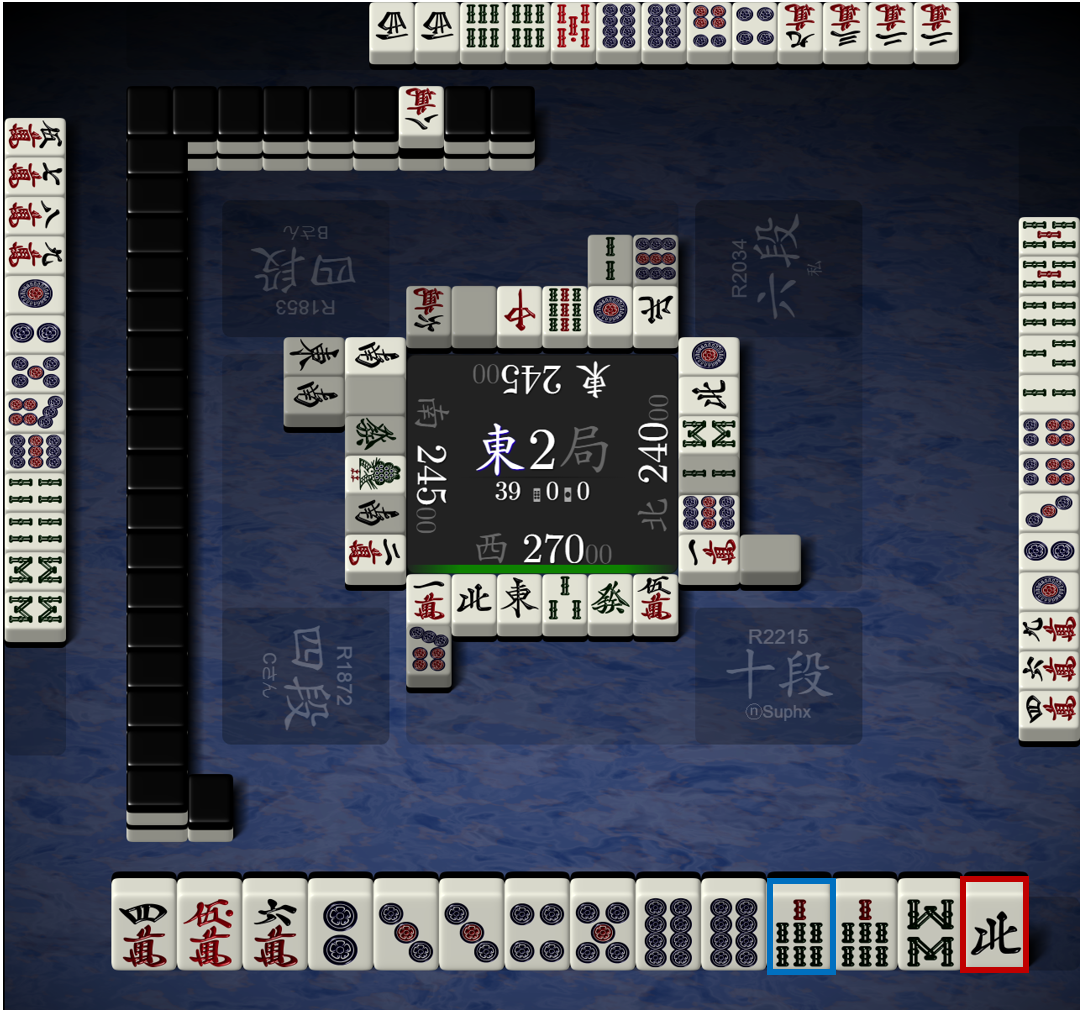}
  \end{center}
  \caption{{Suphx keeps a safe tile at state $s_t$ to balance future attack and defense. Although it is safe to discard the red-boxed tile at state $s_t$ (actually this is indeed the tile that most human players would discard), Suphx keeps this tile in hand; instead, it discards the blue-boxed tile, which might slow down the process of forming a winning hand. Doing so leads to more flexibility in future states and can better balance future attack and defense. Consider a future state $s_{t+k}$ in which another player declares Riichi that is unexpected to our agent. In this case, Suphx can discard the safe tile kept at state $s_t$ and does not break the winning hand it tries to form. In contrast, if Suphx discards this safe tile at state $s_t$, it has no other safe tiles to discard at $s_{t+k}$, and thus may have to break a meld or pair in its hand tiles that are close to a winning hand, consequently resulting in smaller winning probability.  }
  }
  \label{fig:keep-safe-tile}
\end{figure}

Suphx has developed its own playing styles, which are well recognized by top human players. For example, Suphx is very good at keeping safe tiles, prefers winning hand with half-flush\footnote{\url{https://en.wikipedia.org/wiki/Japanese_Mahjong_yaku}}, etc. Figure \ref{fig:keep-safe-tile} is an example that Suphx keeps a safe tile to balance future attack and defense\footnote{Game replay can be found at \url{https://tenhou.net/3/?log=2019070722gm-0029-0000-3bee4a7e&tw=3}.}.

\section{Conclusion and Discussions}
Suphx is the strongest Mahjong AI system up to date, and is also the first  Mahjong AI that surpassed most top human players in Tenhou.net, a famous online platform for Japanese Mahjong. Because of the complexity and unique challenges of Mahjong, we believe that even though Suphx has performed very well, there is still a lot of space for further improvement. 
\begin{itemize}
    \item We introduced global reward prediction in Suphx. In the current system, the reward predictor takes limited information as its input. Clearly, more information will lead to better reward signal. For example, if a \round is very easy to win due to our good luck of initial hand tiles, winning this \round does not reflect the superiority of our policy and should not be rewarded too much; in contrast, winning a difficult \round should be  rewarded more. That is, game difficulty should be taken into consideration while designing reward signals. We are investing how to leverage prefect information (e.g., by comparing the private initial hands of different players) to measure the difficulty of a round/game and then boost the reward predictor. 
    \item We introduced the concept of oracle guiding, and instantiated this concept using the gradual transition from an oracle agent to a normal agent by means of perfect feature dropout. In addition to this, there could be other approaches to leverage the perfect information. For one example, we can simultaneously train an oracle agent and a normal agent, let the oracle agent distill its knowledge to the normal agent while constraining the distance between these two agents. According to our preliminary experiments, this approach also works quite well. For another example, we can consider designing an oracle critic, which provides more effective state-level instant feedback (instead of round-level feedback) to accelerate the training of the policy function based on the perfect information. 
    \item For run-time policy adaptation, in the current system of Suphx, we did simulations at the beginning of each \round when private tiles were dealt to our agent. Actually we can also do simulations after each tile is discarded by any player. That is, instead of only adapting the policy to the initial hands, we can continue the adaptation as the game goes on and more and more information becomes observable. Doing so should be able to further improve the performance of our policy. Moreover, since we gradually adapt our policy, we do not need too many samplings and roll-outs at each step. In other words, we can amortize the computational complexity of policy adaptation over the entire round. With this, it is even possible to use policy adaption in the online playing with affordable computational resources.  
\end{itemize}

Suphx is an agent that constantly learns and improves. Today’s achievement of Suphx on Tenhou.net is just the beginning. Looking forward, we will introduce more novel technologies to Suphx, and continue to push the frontier of Mahjong AI and imperfect-information game playing. 

Most real world problems such as finance market predication and logistic optimization share the same characteristics with Mahjong rather than Go/chess - complex operation/reward rules, imperfect information, etc. We believe our techniques designed in Suphx for Mahjong, including global reward prediction, oracle guiding and parametric Monte-Carlo policy adaptation, have a great potential to benefit for a wide range of real-world applications.

\section*{Acknowledgement}

    We would like to sincerely thank Tsunoda Shungo and Tenhou.net for providing the expert game playing logs and online platform for our experiments. We would like to thank players on Tenhou.net for playing games with Suphx. We would like to thank MoYuan for helping us to collect the statistics of human professional players.  We would also like to thank our interns Hao Zheng and Xiaohong Ji, as well as our colleagues Yatao Li, Wei Cao, Weidong Ma, and Di He for their contributions to developing the learning algorithms and training system of Suphx.

\bibliography{main}

\begin{thebibliography}{10}

\bibitem{berner2019dota}
Christopher Berner, Greg Brockman, Brooke Chan, Vicki Cheung, Przemys{\l}aw
  D{\k{e}}biak, Christy Dennison, David Farhi, Quirin Fischer, Shariq Hashme,
  Chris Hesse, et~al.
\newblock Dota 2 with large scale deep reinforcement learning.
\newblock {\em arXiv preprint arXiv:1912.06680}, 2019.

\bibitem{Bowling2017-tm}
Michael Bowling, Neil Burch, Michael Johanson, and Oskari Tammelin.
\newblock Heads-up limit {Hold'Em} poker is solved.
\newblock {\em Commun. ACM}, 60(11):81--88, October 2017.

\bibitem{Brown2018-wg}
Noam Brown and Tuomas Sandholm.
\newblock Superhuman {AI} for heads-up no-limit poker: Libratus beats top
  professionals.
\newblock {\em Science}, 359(6374):418--424, January 2018.

\bibitem{brown2019superhuman}
Noam Brown and Tuomas Sandholm.
\newblock Superhuman ai for multiplayer poker.
\newblock {\em Science}, 365(6456):885--890, 2019.

\bibitem{European_Mahjong_Association_undated-vi}
{European Mahjong Association}.
\newblock {Rules for Japanese Mahjong}.
\newblock
  \url{http://mahjong-europe.org/portal/images/docs/Riichi-rules-2016-EN.pdf}.

\bibitem{gao2018supervised}
Shiqi Gao, Fuminori Okuya, Yoshihiro Kawahara, and Yoshimasa Tsuruoka.
\newblock Supervised learning of imperfect information data in the game of
  mahjong via deep convolutional neural networks.
\newblock {\em Information Processing Society of Japan}, 2018.

\bibitem{Gao2019-wl}
Shiqi Gao, Fuminori Okuya, Yoshihiro Kawahara, and Yoshimasa Tsuruoka.
\newblock Building a computer mahjong player via deep convolutional neural
  networks.
\newblock June 2019.

\bibitem{Kurita2019-cd}
Moyuru Kurita and Kunihito Hoki.
\newblock Method for constructing artificial intelligence player with
  abstraction to markov decision processes in multiplayer game of mahjong.
\newblock April 2019.

\bibitem{Mizukami2015-vm}
N~Mizukami and Y~Tsuruoka.
\newblock Building a computer mahjong player based on monte carlo simulation
  and opponent models.
\newblock In {\em 2015 {IEEE} Conference on Computational Intelligence and
  Games ({CIG})}, pages 275--283, August 2015.

\bibitem{mizukami2015building}
Naoki Mizukami and Yoshimasa Tsuruoka.
\newblock Building a computer mahjong player based on monte carlo simulation
  and opponent models.
\newblock In {\em 2015 IEEE Conference on Computational Intelligence and Games
  (CIG)}, pages 275--283. IEEE, 2015.

\bibitem{Moravcik2017-cf}
Matej Morav{\v c}{\'\i}k, Martin Schmid, Neil Burch, Viliam Lis{\'y}, Dustin
  Morrill, Nolan Bard, Trevor Davis, Kevin Waugh, Michael Johanson, and Michael
  Bowling.
\newblock {DeepStack}: Expert-level artificial intelligence in heads-up
  no-limit poker.
\newblock {\em Science}, 356(6337):508--513, May 2017.

\bibitem{rong2019competitive}
Jiang Rong, Tao Qin, and Bo~An.
\newblock Competitive bridge bidding with deep neural networks.
\newblock In {\em Proceedings of the 18th International Conference on
  Autonomous Agents and MultiAgent Systems}, pages 16--24. International
  Foundation for Autonomous Agents and Multiagent Systems, 2019.

\bibitem{Silver2016-ws}
David Silver, Aja Huang, Chris~J Maddison, Arthur Guez, Laurent Sifre, George
  van~den Driessche, Julian Schrittwieser, Ioannis Antonoglou, Veda
  Panneershelvam, Marc Lanctot, Sander Dieleman, Dominik Grewe, John Nham, Nal
  Kalchbrenner, Ilya Sutskever, Timothy Lillicrap, Madeleine Leach, Koray
  Kavukcuoglu, Thore Graepel, and Demis Hassabis.
\newblock Mastering the game of go with deep neural networks and tree search.
\newblock {\em Nature}, 529(7587):484--489, January 2016.

\bibitem{silver2016mastering}
David Silver, Aja Huang, Chris~J Maddison, Arthur Guez, Laurent Sifre, George
  Van Den~Driessche, Julian Schrittwieser, Ioannis Antonoglou, Veda
  Panneershelvam, Marc Lanctot, et~al.
\newblock Mastering the game of go with deep neural networks and tree search.
\newblock {\em nature}, 529(7587):484, 2016.

\bibitem{Silver2018-rs}
David Silver, Thomas Hubert, Julian Schrittwieser, Ioannis Antonoglou, Matthew
  Lai, Arthur Guez, Marc Lanctot, Laurent Sifre, Dharshan Kumaran, Thore
  Graepel, Timothy Lillicrap, Karen Simonyan, and Demis Hassabis.
\newblock A general reinforcement learning algorithm that masters chess, shogi,
  and go through self-play.
\newblock {\em Science}, 362(6419):1140--1144, December 2018.

\bibitem{Silver2017-gx}
David Silver, Julian Schrittwieser, Karen Simonyan, Ioannis Antonoglou, Aja
  Huang, Arthur Guez, Thomas Hubert, Lucas Baker, Matthew Lai, Adrian Bolton,
  Yutian Chen, Timothy Lillicrap, Fan Hui, Laurent Sifre, George van~den
  Driessche, Thore Graepel, and Demis Hassabis.
\newblock Mastering the game of go without human knowledge.
\newblock {\em Nature}, 550(7676):354--359, October 2017.

\bibitem{sutton2000policy}
Richard~S Sutton, David~A McAllester, Satinder~P Singh, and Yishay Mansour.
\newblock Policy gradient methods for reinforcement learning with function
  approximation.
\newblock In {\em Advances in neural information processing systems}, pages
  1057--1063, 2000.

\bibitem{Tesauro1995-za}
Gerald Tesauro.
\newblock Temporal difference learning and {TD-Gammon}.
\newblock {\em Commun. ACM}, 38(3):58--68, 1995.

\bibitem{Tsunoda_undated-fr}
Shingo Tsunoda.
\newblock Tenhou.
\newblock \url{https://tenhou.net/}.
\newblock Accessed: 2019-6-17.

\bibitem{Village_undated-jt}
Dwango~Media Village.
\newblock {NAGA}: Deep learning mahjong {AI}.
\newblock \url{https://dmv.nico/ja/articles/mahjong_ai_naga/}.
\newblock Accessed: 2019-6-29.

\bibitem{vinyals2019grandmaster}
Oriol Vinyals, Igor Babuschkin, Wojciech~M Czarnecki, Micha{\"e}l Mathieu,
  Andrew Dudzik, Junyoung Chung, David~H Choi, Richard Powell, Timo Ewalds,
  Petko Georgiev, et~al.
\newblock Grandmaster level in starcraft ii using multi-agent reinforcement
  learning.
\newblock {\em Nature}, 575(7782):350--354, 2019.

\end{thebibliography}
\bibliographystyle{plain}

\newpage
\appendix

\section*{Appendix A: Rules of Mahjong}
\label{app:rule_of_mahjong}
Mahjong is a tile-based game that was developed in China hundreds of years ago and is now popular worldwide with hundreds of millions of players. The game of Mahjong itself has numerous variations across the world, and Mahjong in different regions is different in both the rules and the tiles used. In this work, we focus on 4-player { Japanese Mahjong (Riichi Mahjong)} considering that Japanese Mahjong is very popular with a professional league\footnote{\url{https://m-league.jp/}} of top players in Japan and its rules~\cite{European_Mahjong_Association_undated-vi} are clearly defined and well accepted. Since the playing/scoring rules of Japanese Mahjong are very complex and the focus of this work is not to give a comprehensive introduction to them, here we make some selections and simplifications and give a brief introduction to those rules. Comprehensive introductions can be found at Mahjong International League rules ~\cite{European_Mahjong_Association_undated-vi}.

There are 136 tiles in Japanese Mahjong, consisting of 34 different kinds of tiles, with four of each kind. The 34 tiles are consists of three suit, {Bamboo}, {Character} and {Dot}, each from 1 to 9, and 7 different {Honour} tiles. A game contains multiple rounds and ends when one player loses all points, or some winning condition is triggered.

Each player in a game will start with 25,000 points, and one of four players is designated as the dealer. At the beginning of each round, all the tiles are shuffled and arranged into four walls, each with 34 tiles. 52 tiles are dealt to 4 players (13 for each player as his/her private hand), 14 tiles form the \textit{dead wall}, which are never played except when players declare Kongs and draw a replacement tile, and the remaining 70 tiles form the live wall. 
The 4 players take turns to draw and discard tiles.\footnote{The turns can be interrupted by a Kong/Kong and declaring a winning hand.} The player who first builds a \textit{complete hand} with at least one {\yaku} wins the round and gets certain round score calculated by the rewarding rules:
\begin{itemize}
    \item A {complete hand} is a set of 4 \textit{melds} plus a \textit{pair}. A {meld} can be a \textit{Pong} (three identical tiles), a \textit{Kong} (four identical tiles), and a \textit{Chow} (three consecutive simple tiles of the same suit). The {pair} consists of any two identical tiles, but it cannot be mixed with the four {melds}. A player can make a Chow/Pong/Kong from (1) a tile drawn from the wall by himself/herself, in which case this meld is concealed to others, or (2) a tile discarded by other players, in which case this meld is exposed to others. If a Kong is made from a tile drawn from the wall, it is called \textit{ClosedKong}. After a player makes a {Kong}, he/she needs to draw an additional tile from the {dead wall} for replacement. A special case called \textit{AddKong} converts an exposed {Pong} into a Kong when a player draws a tile that matches an exposed \textit{Pong} he/she has.

\item A {\yaku} is a certain pattern of players' tiles or a special condition. {Yaku} is the main factor to determine round score, and the value of \yaku varies from different patterns. A winning hand could contain several different \yaku's and the final round score will be accumulated across all the \yaku's in hand. Different variations of Japanese Majong have different \yaku patterns. A common list of \yaku consist of {40} different types. Furthermore, \dora, a special tile determined by rolling dice before drawing tiles, provides additional points as a bonus. 
\end{itemize}

The last tile that a player forms a winning hand together with his/her private tiles can come from (1) a tile from the wall drawn by himself/herself or (2) a tile discarded by other players. For the first case, all other players will lose points to the winner. For the second case, the player who discards the tile will lose points to the winner.

A special \yaku is that player can declare Riichi when his/her hand is only one tile away from a winning hand. Once declaring a Riichi, the player can only claim winning hand from either a self-drawn tile or  a tile discarded by other players, and he/she is not allowed to change his/her hand tiles any more.\footnote{\url{http://mahjong.wikidot.com/riichi}}

The final ranking points a player gets from a game is determined by his/her level and the rank of his/her accumulated round score over the multiple rounds of the game, as shown in Table \ref{table:tenhou_rank_system}.

\section*{Appendix B: Tenhou Ranking Rules}
\label{app:tenhou_house_rule}

\begin{table}[]
\centering
\resizebox{\textwidth}{!}{%
\begin{tabular}{@{}cccccccc@{}}
\toprule
Level                        & \begin{tabular}[c]{@{}c@{}}Base\\ Ranking Pts.\end{tabular} & 1st & 2nd & 3rd & 4th  & \begin{tabular}[c]{@{}c@{}}Level Up\\ Ranking Pts.\end{tabular} & Level Down \\ \midrule
Rookie &
  0 &
  \multirow{20}{*}{\begin{tabular}[c]{@{}c@{}}+20 \\ @Normal Room \\ \\ \\ \\ +40\\ @Advanced Room \\ \\ \\ \\ +50\\ @Expert Room \\ \\ \\ \\ +60\\ @Phoenix Room\end{tabular}} &
  \multirow{20}{*}{\begin{tabular}[c]{@{}c@{}}+10 \\ @Normal Room \\ \\ \\ \\ +10\\ @Advanced Room \\ \\ \\ \\ +20\\ @Expert Room \\ \\ \\ \\ +30\\ @Phoenix Room\end{tabular}} &
  \multirow{20}{*}{+0} &
  0 &
  20 &
  - \\
9Kyu                        & 0        &     &     &     & 0    & 20                                                      & -          \\
8Kyu                        & 0        &     &     &     & 0    & 20                                                      & -          \\
7Kyu                        & 0        &     &     &     & 0    & 20                                                      & -          \\
6Kyu                        & 0        &     &     &     & 0    & 40                                                      & -          \\
5Kyu                        & 0        &     &     &     & 0    & 60                                                      & -          \\
4Kyu                        & 0        &     &     &     & 0    & 80                                                      & -          \\
3Kyu                        & 0        &     &     &     & 0    & 100                                                     & -          \\
2Kyu                        & 0        &     &     &     & -15  & 100                                                     & -          \\
1Kyu                        & 0        &     &     &     & -30  & 100                                                     & -          \\
1Dan                        & 200      &     &     &     & -45  & 400                                                     & Yes        \\
2Dan                        & 400      &     &     &     & -60  & 800                                                     & Yes        \\
3Dan                        & 600      &     &     &     & -75  & 1200                                                    & Yes        \\
4Dan                        & 800      &     &     &     & -90  & 1600                                                    & Yes        \\
5Dan                        & 1000     &     &     &     & -105 & 2000                                                    & Yes        \\
6Dan                        & 1200     &     &     &     & -120 & 2400                                                    & Yes        \\
7Dan                        & 1400     &     &     &     & -135 & 2800                                                    & Yes        \\
8Dan                        & 1600     &     &     &     & -150 & 3200                                                    & Yes        \\
9Dan                        & 1800     &     &     &     & -165 & 3600                                                    & Yes        \\
10Dan                       & 2000     &     &     &     & -180 & 4000                                                    & Yes        \\
\multicolumn{1}{l}{Phoenix} & \multicolumn{6}{c}{Honorable Title}                                                         & -          \\ \bottomrule
\end{tabular}%
}
\caption{{Tenhou ranking systems: different levels and their requirements}}
\label{table:tenhou_rank_system}
\end{table}

Tenhou uses the Japanese Martial Arts Ranking System\footnote{\url{https://en.wikipedia.org/wiki/Dan_(rank)}}, which starts from {rookie}, {9 kyuu} down to {1 kyuu}, and then {1 dan} up to {10 dan}. Players earn/lose ranking points when they win/lose ranked games. The amount earned or lost depends on the game result (which ranges from 1 to 4), the current level of the player, and the room that the player is in. There are four types of room in Tenhou, {normal room}, {advanced room}, {expert room} and {phoenix room}. The punishment (i.e., the negative ranking points) of losing a game is the same across these rooms, but the reward (i.e., the positive ranking points) of winning a game are different. The ranking system is designed to punish high-level players if he/she loses a game: a player of a higher level will lose more point for the 4-th rank of game than that of a lower-level player, e.g., -180 for a 10-dan player vs. -120 for a 6-dan player. 

When a player plays a game, he/she wins/loses some ranking points from the game, and his/her total ranking points change. If the total ranking points increase and reach the requirement of the next level, his/her rank increases 1 level; if the total ranking points decrease to 0, his/her rank decreases 1 level. When his/her rank level changes, he/she will get {initial ranking points} at the new level. The details of the ranking points across different rooms and levels are listed in Table \ref{table:tenhou_rank_system}. Therefore, the rank of a player is not stable and often change over time. We use \emph{record rank} to denote the highest rank a player has ever achieved in Tenhou.

\section*{Appendix C: Stable Rank}
Tenhou uses stable rank to evaluate the long-term average performance of a  player. The stable rank in the expert room is calculated as follows.\footnote{\url{https://tenhou.net/man/\#RANKING}} 

Let $n_1$ denote the number of \matches a player gets the highest accumulated \round scores, $n_4$ the number of \matches he/she gets the lowest accumulated \round scores, and $n_2$ and $n_3$ the numbers of \matches for the second/third highest accumulated \round scores. Then the stable rank of the player in terms of dan is 
\begin{equation}
    \label{eq:dan}
    \frac{5\times n_1 + 2\times n_2}{n_4}-2.
\end{equation}

Since the accumulated \round score of a \match depends not only on the skills of the player but also on the private tiles of the four players and the wall tiles, the stable rank could be of large variance due to the randomness in the hidden information. Furthermore, when playing in Tenhou, opponents are randomly allocated by the Tenhou system, which brings in additional randomness. Thus, for a player in Tenhou, it is usually assumed that at least a few thousands of \matches are needed to get a relatively reliable stable rank.

\end{document}